\documentclass[conference]{IEEEtran}
\IEEEoverridecommandlockouts

\usepackage{amsmath,amssymb,amsfonts}
\usepackage{algorithmic}
\usepackage{graphicx}
\usepackage{textcomp}
\usepackage{xcolor}
\usepackage[numbers]{natbib}
\usepackage{todonotes}
\usepackage{siunitx}
\usepackage{orcidlink}
\usepackage{multirow}
\usepackage{booktabs}

\def\BibTeX{{\rm B\kern-.05em{\sc i\kern-.025em b}\kern-.08em
    T\kern-.1667em\lower.7ex\hbox{E}\kern-.125emX}}
\begin{document}

\title{CapsuleMotion: A Lightweight Real-Time Visual Motion Predictor for Capsule Endoscopy

\thanks{$^1$University of T\"ubingen, Faculty of Science, Department of Computer Science, Embedded Systems Group,
\tt\small {[first].[lastname]@uni-tuebingen.de}}

\author{
    \IEEEauthorblockN{Oliver Bause$^1$}
    \IEEEauthorblockA{
    \orcidlinkc{0009-0003-5388-2959}}
    \and
    \IEEEauthorblockN{Julia Werner$^1$}
    \IEEEauthorblockA{
    \orcidlinkc{0009-0006-0279-1776}}
    \and
    \IEEEauthorblockN{Oliver Bringmann$^1$}
    \IEEEauthorblockA{
    \orcidlinkc{0000-0002-1615-507X}}
}
}

\maketitle

\begin{abstract}
Video Capsule Endoscopy (VCE) is a non-invasive medical examination that allows for the observation of the small intestine, which is otherwise difficult to access.
A fundamental challenge persists in the form of their limited size in order to still be swallowable.
The resulting restricted battery capacity, however, contradicts with the power-intensive nature of image capture and transmission.
Therefore, we propose CapsuleMotion, a patient-specific dynamic capsule behavior that utilizes the available energy in a goal-oriented manner to increase the likelihood of a complete screening of the gastrointestinal tract.
By investigating and combining metrics from the on-device image compression, CapsuleMotion predicts the motion between two successive frames.
The camera's frame rate will be modified in accordance with the predicted magnitude of motion.
Furthermore, prior to entering the small intestine, the capsule operates in a low power mode with a significantly reduced frame rate.
In this mode, the LocalizationNet is employed to determine the current organ, provided that motion was predicted.
The proposed framework is evaluated on the Rhode Island VCE dataset and deployed on an ultra-low power single-core RISC-V demonstrator with an integrated hardware accelerator. 
CapsuleMotion demonstrated the capability to reduce electric energy consumption by up to $20.66\%$ in comparison with conventional capsules that lack a dynamic frame rate.
Additionally, the accuracy of detecting the entry point of the small intestine has been improved.
\end{abstract}

\begin{IEEEkeywords}
Video Capsule Endoscopy, On-Device Motion Estimation, Capsule Localization, Image Compression
\end{IEEEkeywords}

\section{Introduction}
Introduced in the early 2000s, wireless Video Capsule Endoscopy (VCE) utilizes a pill-sized device that patients swallow to examine the gastrointestinal (GI) tract~\cite{iddan2000wireless,swain2001wireless}. 
The capsule integrates a miniature image sensor, LED illumination, a transmitter, a battery, and a microcontroller, with the potential to incorporate additional sensors to expand diagnostic capabilities.
As the Video Capsule (VC) travels through the GI tract, it captures images and transmits them to an external on-body receiver for later clinical analysis. 
This technology is primarily used to identify pathologies in the small intestine, a region otherwise inaccessible to standard gastroscopy or colonoscopy~\cite{costamagna2002prospective}.

A significant challenge for VCE is the limited battery life, a direct result of the strict physical constraints required to ensure the capsule remains swallowable.
For instance, the Medtronic PillCam™ SB3 offers an operational window of 8 to 12 hours at a frame rate of 2 to 6 frames per second (fps)~\cite{pillcam}. 
However, as the time required for a capsule to traverse the GI tract varies greatly among patients and can exceed 12 hours, the procedure often risks being incomplete.
This limitation can result in blind spots where potential pathologies in the small intestine or colon remain undetected.

\begin{figure}[t]
    \centering
    \includegraphics[width=\linewidth]{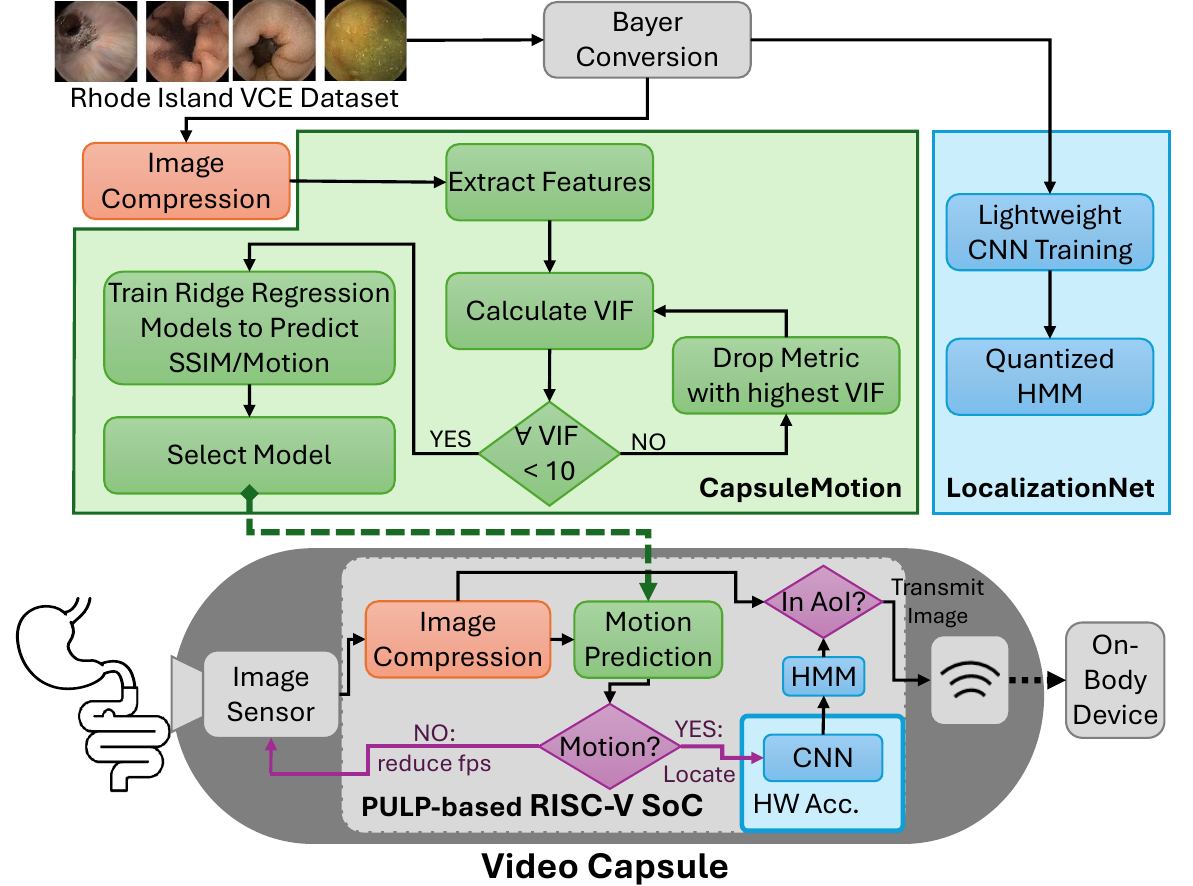}
    \caption{Proposed pipeline to train a motion predictor based on features that can be extracted from the image compression and combining it with a lightweight CNN to achieve a more precise localization and energy efficient screening.}
    \label{fig:pipeline}
\end{figure}
\textbf{Our Contribution:} To increase the probability of a complete screening, this work introduces a patient-specific dynamic system behavior that adjusts the frame rate depending on the current location within the GI tract and the capsule's estimated motion.
The proposed pipeline is illustrated in Figure~\ref{fig:pipeline}.
First, CapsuleMotion investigates metrics from a hardware-suitable Adaptive Golomb-Rice (AGR) coding compression technique~\cite{bause2026imagecompression, rice1971coding} that supports the RAW Bayer output from the image sensor without requiring further pre-processing.
These metrics are then used to predict the Mean Structural Similarity Index Measure (MSSIM)~\cite{wang2004image} of two successive frames with a ridge regression model~\cite{mcdonald2009ridge}.
In the event that the MSSIM falls below a certain threshold, motion is assumed, and the organ where the frame was taken is predicted on-device with a combination of a lightweight Convolutional Neural Network (CNN) and a Hidden Markov Model (HMM) with Viterbi Decoding, termed LocalizationNet~\cite{bause2025smartVCE}, which contains only 64,000 8-bit weights. 
The motion-triggered localization reduces the amount of consecutive mislabeling within a difficult scene.
As a result, this enables the deployment of a restrictive low power mode that drastically reduces the frame rate to save energy and is active until the Area of Interest (AoI), the small intestine, is reached.
After the localization detect the transition into the AoI, the system will be set into the normal operation mode and the frame rate is adjusted according to CapsuleMotion's prediction.
The pipeline has been validated with the Rhode Island (RI) gastroenterology VCE dataset~\cite{charoen2022rhode} and demonstrated on an ultra-low power single-core RISC-V System-on-Chip (SoC)~\cite{bernardo2024scalable} with the hardware accelerator UltraTrail~\cite{palomero2025compiler} integrated.
\section{Related Work}
Visual motion estimation has been widely studied in computer vision through visual odometry and ego-motion estimation. 
Raudies et al. conducted a review of classical methods based on optical flow, feature tracking, and geometric reconstruction, emphasizing the trade-off between accuracy and computational cost~\cite{raudies2012review}.

In the context of VCE, the process of motion estimation is characterized by a heightened degree of complexity, primarily due to tissue deformation, low-texture surfaces, and illumination changes. 
Early work by Liu et al. estimated capsule motion directly from monocular video for redundancy reduction in VCE recordings~\cite{liu2013wireless}. 
Later, Spyrou et al. evaluated feature-based visual odometry methods for capsule localization, showing that classical descriptors such as SIFT and SURF can provide useful motion information despite challenging imaging conditions \cite{spyrou2015comparative}.

However, recent approaches have shifted toward the utilization of deep learning methodologies. 
Turan et al. proposed Deep EndoVO, an RCNN-based framework for end-to-end monocular pose estimation from endoscopic video \cite{turan2018deep}. 
While the effectiveness of these methods is evident, their high computational complexity restricts their applicability to devices with limited resources. 
These methods are intended for implementation subsequent to screening and prior to analysis by medical professionals.
Therefore, their deployment within the VC to execute real-time motion estimation, which directly impacts the device's operation, is not feasible.

 Sensor-based localization approaches, in contrast, utilize onboard accelerometers, gyroscopes, or magnetometers to estimate capsule motion and orientation. 
 Despite their computational efficiency, the sensors require space within the capsule, and their accuracy is limited by several factors. 
 Among these are sensor drift, accumulated integration errors, magnetic interference, and patient movement, which can introduce significant localization uncertainty over time. 
 Consequently, numerous systems mandate the use of external sensors, magnetic tracking hardware, or sensor-fusion techniques to ensure the reliability of pose estimates~\cite{ali2025recent}.
\section{Methodology}

\subsection{Rhode Island VCE Dataset}
The RI VCE dataset~\cite{charoen2022rhode} was used to develop and validate the proposed pipeline.
It consists of 424 complete studies with a total of over five millions frames labeled with their respective anatomical organ in the GI tract.
LocalizationNet was trained using the official downsampled training and validation splits~\cite{bause2025smartVCE}.
However, to train the ridge regression models, the 339 complete studies from the training and validation splits were utilized and randomly divided into an $80/20\%$ split by study.
It was imperative to ensure that the regression model precisely predicted the MSSIM of successive frames, rather than the MSSIM of two random frames from different patients. 
Furthermore, the MSSIM of sequenced frames at varying frame rates was analyzed to verify the assumption that it can be used as a motion predictor. 
As anticipated, the mean MSSIM between successive frames exhibits a decline in value as the frame rate is reduced.
Nonetheless, the evaluation of both models was conducted by utilizing the official test set, which encompasses 85 complete studies.

\subsection{LocalizationNet}
In preceding work \cite{bause2025smartVCE}, we presented the ultra-low power LocalizationNet that is a combination of a lightweight CNN with a quantized HMM requiring only \qty{5.31}{\micro\joule} per inference to precisely determine the organ of the current frame.
Additional hardware-level optimizations further reduced it to just $\qty{1.53}{\micro\joule}$ per inference.
To enhance the image classification results, the CNN was combined with Viterbi decoding which computes the most likely sequence of hidden states given the HMM and the CNN predictions based on \cite{werner2023precise}.
The sequence of hidden states directly corresponds to the presumed path traversed by the capsule through the organs.
Furthermore, by reducing the frame rate, the accuracy of the network was increased which also allowed a reduction of the HMM window size and resulted in an overall power reduction before entering the AoI.
The improvement in accuracy can be attributed to a reduction in the number of sequential frames analyzed from scenes that were challenging for the network to predict, leading to a decrease in consecutive mislabelings.

\subsection{Bayer Compression Pipeline}
The efficient AGR compression pipeline, demonstrated in \cite{bause2026imagecompression}, achieved great compression results and, thus, was adapted for this framework.
The compression was validated on the RI dataset and achieved similar results with an average compression ratio (CR) of $3.99$ and a Peak Signal-to-Noise Ratio (PSNR) of \qty{38.88}{\dB}.
It was then adjusted to also log different metrics during compression which were utilized to train the ridge regression MSSIM prediction model.

\subsection{CapsuleMotion}
To estimate the motion of the capsule, we introduce the CapsuleMotion pipeline as illustrated in Figure~\ref{fig:pipeline}.
The goal is to predict the MSSIM between two sequential frames.
Conversely, if the MSSIM exceeds $0.95$, the images are deemed to be so similar that the motion between these two frames is considered negligible. 
Consequently, the captured image is deemed obsolete and is not transmitted to the on-body receiver.
If it is higher than $0.85$, the motion is slow and the capsule's frame rate can be reduced to save power.
Otherwise, the device operates at the pre-defined normal frame rate, typically between 2 and 6 fps.
However, calculating the MSSIM on-device is too computationally complex and energy intensive for an ultra-low power SoC which would evaporate any possible energy and time savings.
Thus, the CapsuleMotion pipeline trains a simple ridge regression model that predicts the MSSIM between two frames with a few selected metrics that can be easily and efficiently extracted during the compression of a frame.

The pipeline starts by identifying possible metrics which can be extracted during the image compression.
By compressing the images of the RI's train set, $37$ different metrics were identified:
\begin{itemize}
    \item Sum of Absolute Difference (SAD) of each DCT component between two frames (4),
    \item Hamming Distance (HD) and length difference within the three encoded bitstrings (6),
    \item CR and PSNR (2),
    \item and SAD between frame slices by slicing the image into a $5\times5$ grid (25).
\end{itemize}

Then, an iterative feature selection based on the Variance Inflation Factor (VIF) eliminates stepwise metrics to mitigate multicollinearity in regression models. 
In each iteration, the algorithm computes the $\text{VIF}_j$ for each remaining independent metric $X_j$ by first constructing an auxiliary ordinary least squares (OLS) regression. 
In this auxiliary model, $X_j$ is treated as the dependent variable and regressed against all other independent variables in the current subset:
\begin{equation}
X_j = \beta_0 + \sum_{k \neq j} \beta_k X_k + \epsilon.
\end{equation}
From this auxiliary regression, the coefficient of determination $R_j^2$ is extracted, which quantifies the proportion of variance in $X_j$ that is explained by the other predictors. The $\text{VIF}_j$ for variable $X_j$ is then calculated as:
\begin{equation}
\text{VIF}_j = \frac{1}{1 - R_j^2}.
\end{equation}
The variable that exhibits the highest VIF is identified. 
In the event that the maximum value exceeds a threshold of 10, as this is regarded as high~\cite{sheather2009modern} , the corresponding metric is eliminated from the feature subset, and the process is repeated.
This cycle continues iteratively until the $\text{VIF}_j$ values of all remaining metrics fall below the specified threshold.
Subsequently, ridge regression models were trained and validated on the validation set, incorporating all potential metric combinations of up to 10 features. 
This process resulted in the development of over 4,000 trained models. 
The objective was to select the best-performing one per number of features, which could then deployed into the VC.
\section{Results}
\subsection{Feature Selection}
\begin{figure}[t]
    \centering
    \includegraphics[width=\linewidth]{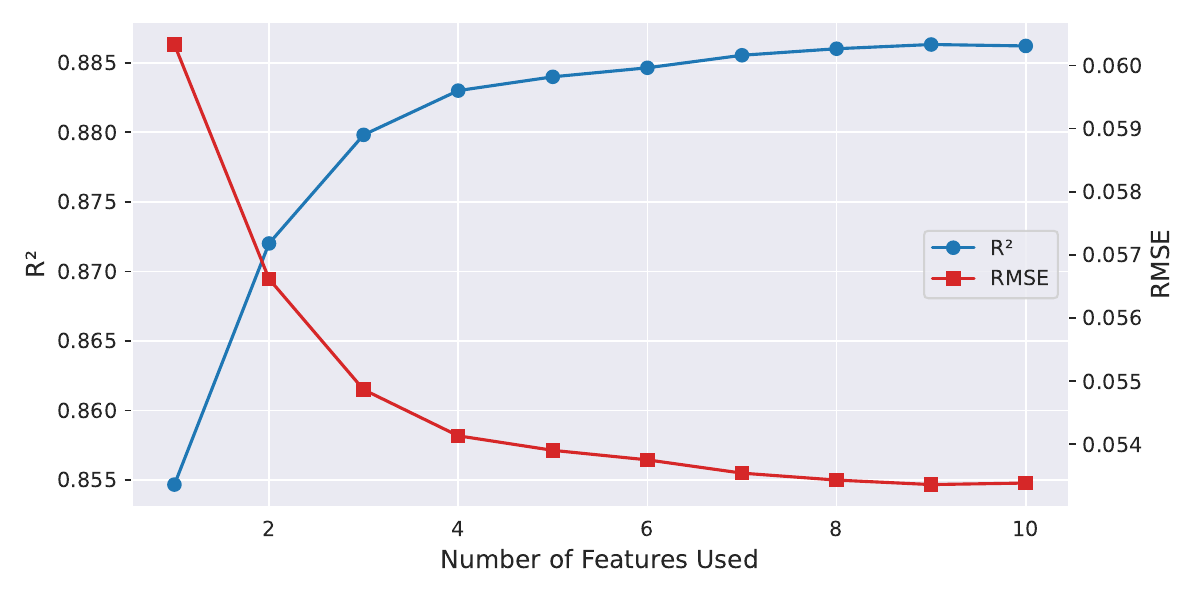}
    \caption{SSIM predictor performance over different number of metrics utilized.}
    \label{fig:ssim_predictor_performance}
\end{figure}

\begin{table}[t]
    \centering
    \setlength{\tabcolsep}{0.5em}
    \caption{Metrics used in the four smallest ridge regression models and their corresponding inference time to compute them.}
    \begin{tabular}{l | *{4}{w{c}{2em}} | c}
        \toprule
        \multirow{2}{*}{\textbf{Metric}}  & \multicolumn{4}{c|}{\textbf{Used in Model}} & \textbf{Inference} \\
        & \textbf{1} & \textbf{2} & \textbf{3} & \textbf{4} & \textbf{Time} [$\unit{\ms}$] \\
        \midrule
        SAD of AGR coefficient \textit{y2} & x & x & x & x & $1.792$ \\
        HD of encoded \textbf{y} bitstring & - & x & - & x & $0.0003$ \\
        SAD of AGR coefficient \texttt{cr} & - & - & x & x & $1.764$ \\
        SAD of $64\times64$px center crop & - & - & x & x & $0.279$ \\
        \midrule
        RISC-V optimized SSIM & - & - & - & - & $26.41$ \\
        \midrule
        \midrule
        Total Model Inference Time [$\unit{\ms}$] & $1.795$ & $1.799$ & $3.844$ & $3.848$ & \\
        \bottomrule
    \end{tabular}
    \label{tab:metrics}
\end{table}
After the iterative VIF-based feature elimination, 12 metrics remained in the set for possible ridge regression models.
For each number of features, the best performing model on the validation set was selected and tested against the RI test set.
The achieved coefficient of determination (R$^2$) and Root Mean Square Error (RMSE) of these 10 models are illustrated in Figure~\ref{fig:ssim_predictor_performance}.
The regression model demonstrates minimal enhancement in accuracy when more than four features are employed. 
Given that each additional feature increases the computational load, the analysis of the hardware demonstrator has been restricted to the four smallest models which are listed with their utilized features in Table~\ref{tab:metrics}.
However, even the calculation of the 4 metrics plus the inference of model 4 requires with $\qty{3.85}{\ms}$ significantly less time than the calculation of the SSIM with $\qty{26.41}{\ms}$ directly.

\subsection{Hardware Demonstrator}
\begin{table}[t]
\centering
\setlength{\tabcolsep}{0.5em}
\caption{Power and electric energy consumption of the demonstrator's modules}
\scriptsize
\begin{tabular}{c|c|ccc}
\toprule
\textbf{Task} & \textbf{Module} & \textbf{IT} [$\unit{\ms}$] & \textbf{Power} [$\unit{\mW}$] & \textbf{Energy} [$\unit{\uJ}$] \\
\midrule
\multirow{3}{*}{\begin{tabular}{c}
Image\\
Capture
\end{tabular}} & NanEyeC & $12.79$ & $8.51$ & $108.93$ \\
& LEDs & $12.79$ & $14.78$ & $189.15$ \\
& Core & $12.79$ & $1.06$ & $13.56$ \\
\midrule
LocalizationNet & Core + Acc. & $0.33$ & $4.62$ & $1.53$ \\
\midrule
Compression & Core & $51.8-94.7$ & $4.28$ & $221.7-405.3$ \\
\midrule
CapsuleMotion & Core & $1.8-3.89$ & $4.28$ & $7.72-16.65$ \\
\midrule
Transmission & Transceiver & $7.5-51.2$ & $9.2$ & $69-475.1
$\\
\midrule
Idle & System & - & $0.43$ & -\\
\bottomrule
\end{tabular}
\label{tab:demonstrator}
\end{table}
The demonstrator was equipped with the $320\times320$px NanEyeC camera, four LEDs, a transceiver, and a single-core RISC-V PULPissimo SoC~\cite{bernardo2024scalable}.
The SoC's die size is just $1.8\times2\unit{\mm}^2$ and has the ultra-low power hardware accelerator UltraTrail integrated which executes the CNN part of the LocalizationNet.
During compression and inference, the core was clocked at $\qty{200}{\MHz}$, while it was reduced to $\qty{25}{\MHz}$ during image capturing and idle times to save power.
The resulting electric energy consumptions split by module are listed in Table~\ref{tab:demonstrator}.

\subsection{Test Study Simulation}
\begin{figure}[t]
    \centering
    \includegraphics[width=\linewidth]{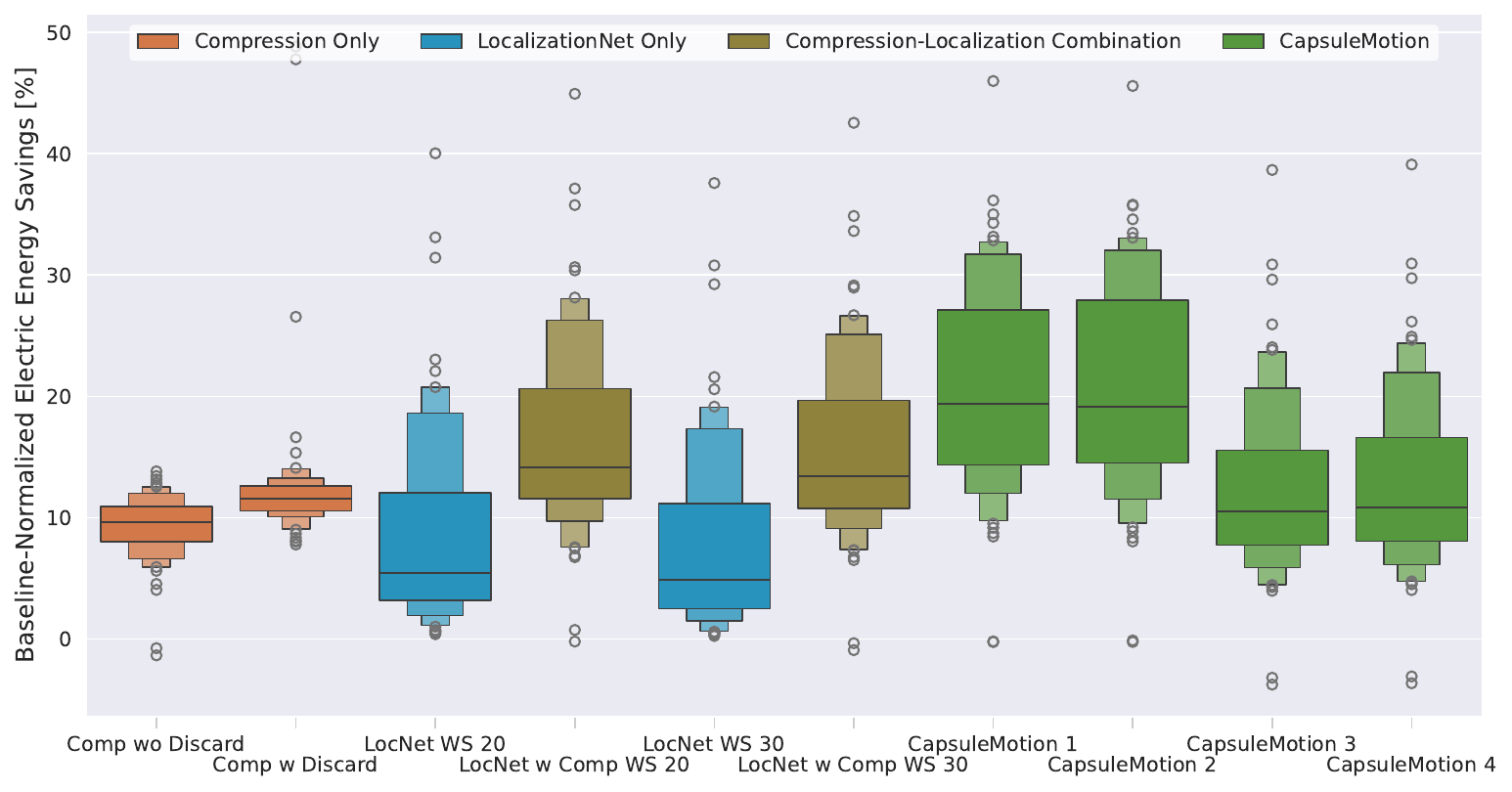}
    \caption{Electric energy savings achieved by the compression~\cite{bause2026imagecompression}, LocalizationNet~\cite{bause2025smartVCE} with a HMM window size of 20 and 30 at a reduced fps of 0.25 and 0.5, respectively, the combination of LocalizationNet and compression, and the presented CapsuleMotion models 1 to 4 normalized against a baseline VCE.}
    \label{fig:energy_per_model}
\end{figure}
All 85 test studies from the RI dataset were emulated on the demonstrator to validate its performance.
Various operation modes were executed and compared against a baseline VC mode that captures images at a constant frame rate of $2$ fps and directly transmit them to the on-body receiver without further analysis.
Figure~\ref{fig:energy_per_model} illustrates the achieved electric energy savings across the different modes in comparison to the baseline.
While the image compression~\cite{bause2026imagecompression} and the LocalizationNet~\cite{bause2025smartVCE} stand-alone achieved an average reduction by $10.90\%$ and $8.01\%$, respectively, their combination even reached up to $16.01\%$.
CapsuleMotion, however, further pushed the average savings to $20.47\%$ and $20.66\%$ for the regression models 1 and 2, respectively, while using only a Viterbi decoding window size of 5 instead of the 20 or 30 of the LocalizationNet modes.
As model 3 and 4 require $114.2\%$ more computational time than the smaller models, the energy savings achieved by the dynamic frame rate are canceled out by the more complex models, resulting in average savings of just $12.3\%$. 
Besides the power savings, the models needs to detect the entry into the AoI precisely.
A late detection risks blind spots in the beginning of the small bowel while an early detection wastes battery by capturing and transmitting stomach images.
The LocalizationNets with a Viterbi window size of $20$ and $30$ achieved a median detection delay of $86.5$ and $\qty{65.5}{\s}$, respectively.
This is still within the area of the small bowel, that is accessible by the classic gastroscopy.
However, all CapsuleMotion models reduced the median delay across all models to an early detection at $\qty{-17.5}{\s}$.
\section{Conclusion}
This work presents the efficient ridge regression-based motion estimator CapsuleMotion.
It enables a patient-specific dynamic VCE screening without the need of additional onboard accelerometers or gyroscopes by extracting existing metrics from the compression pipeline and using them to predict the capsule's motion.
In combination with the lightweight LocalizationNet, the average electric energy savings per screening reached up to $20.66\%$ compared to a standard capsule without a dynamic frame rate.
At the same time, the median detection delay of the small bowel was reduced from more than $60$ seconds late to just $\qty{17.5}{\s}$ early.
For future research, we want to validate this approach on further datasets and evaluate other possible motion predictors based on additional sensors, sensor fusions, or lightweight CNNs.

\bibliographystyle{IEEEtran}
\scriptsize
\bibliography{paper.bib}

@article{iddan2000wireless,
  title={Wireless capsule endoscopy},
  author={Iddan, Gavriel and Meron, Gavriel and Glukhovsky, Arkady and Swain, Paul},
  journal={Nature},
  volume={405},
  number={6785},
  pages={417--417},
  year={2000},
  publisher={Nature Publishing Group UK London}
}

@inproceedings{swain2001wireless,
  title={Wireless capsule endoscopy of the small bowel: development, testing, and first human trials},
  author={Swain, Paul and Iddan, Gavriel J and Meron, Gavriel and Glukhovsky, Arkady},
  booktitle={Biomonitoring and Endoscopy Technologies},
  volume={4158},
  pages={19--23},
  year={2001},
  organization={SPIE}
}

@article{costamagna2002prospective,
  title={A prospective trial comparing small bowel radiographs and video capsule endoscopy for suspected small bowel disease},
  author={Costamagna et al., Guido},
  journal={Gastroenterology},
  volume={123},
  number={4},
  pages={999--1005},
  year={2002},
  publisher={Elsevier}
}

@misc{pillcam,
  author = {Medtronic},
  title = {{PillCam™ SB3 System}},
  howpublished = "\url{https://www.medtronic.com/covidien/en-nz/products/capsule-endoscopy/pillcam-sb-3-system.html/}",
  year = {2025}, 
  note = "[Online; accessed 7-May-2025]"
}

@ARTICLE{rice1971coding,
  author={Rice, R. and Plaunt, J.},
  journal={IEEE Transactions on Communication Technology}, 
  title={Adaptive Variable-Length Coding for Efficient Compression of Spacecraft Television Data}, 
  year={1971},
  volume={19},
  number={6},
  pages={889-897},
  doi={10.1109/TCOM.1971.1090789}
}

@InProceedings{bause2025smartVCE,
author="Bause, Oliver
and Werner, Julia
and Palomero Bernardo, Paul
and Bringmann, Oliver",
title="Smart Video Capsule Endoscopy: Raw Image-Based Localization for Enhanced GI Tract Investigation",
booktitle="Neural Information Processing",
year="2026",
publisher="Springer Nature Singapore",
address="Singapore",
pages="33--47",
isbn="978-981-95-4378-6"
}

@inproceedings{bernardo2024scalable,
  title={A scalable risc-v hardware platform for intelligent sensor processing},
  author={Bernardo et al., Paul Palomero},
  booktitle={2024 Design, Automation \& Test in Europe Conference \& Exhibition (DATE)},
  pages={1--5},
  year={2024},
  organization={IEEE}
}

@inproceedings{bause2026imagecompression,
  title     = {Image Compression with Bubble-Aware Frame Rate Adaptation for Energy-Efficient Video Capsule Endoscopy},
  author    = {Oliver Bause and J{\"o}rg Gamerdinger and Julia Werner and Oliver Bringmann},
  booktitle = {Proceedings of the 48th Annual International Conference of the IEEE Engineering in Medicine and Biology Society (EMBC)},
  year      = {2026},
  note      = {Accepted for publication},
  url       = {https://arxiv.org/abs/2604.25464}
}

@inproceedings{palomero2025compiler,
  title={Compiler-aware AI Hardware Design for Edge Devices},
  author={Palomero Bernardo, Paul and Schmid, Patrick and Gerum, Christoph and Bringmann, Oliver},
  booktitle={Proceedings of the 8th International Workshop on Edge Systems, Analytics and Networking},
  pages={31--36},
  year={2025}
}

@article{charoen2022rhode,
  title={Rhode Island gastroenterology video capsule endoscopy data set},
  author={Charoen, Amber and Guo, Averill and Fangsaard, Panisara and Taweechainaruemitr, Supakorn and Wiwatwattana, Nuwee and Charoenpong, Theekapun and Rich, Harlan G},
  journal={Scientific Data},
  volume={9},
  number={1},
  pages={602},
  year={2022},
  publisher={Nature Publishing Group UK London}
}

@article{raudies2012review,
  title={A review and evaluation of methods estimating ego-motion},
  author={Raudies, Florian and Neumann, Heiko},
  journal={Computer Vision and Image Understanding},
  volume={116},
  number={5},
  pages={606--633},
  year={2012},
  publisher={Elsevier}
}

@article{turan2018deep,
  title={Deep endovo: A recurrent convolutional neural network (rcnn) based visual odometry approach for endoscopic capsule robots},
  author={Turan, Mehmet and Almalioglu, Yasin and Araujo, Helder and Konukoglu, Ender and Sitti, Metin},
  journal={Neurocomputing},
  volume={275},
  pages={1861--1870},
  year={2018},
  publisher={Elsevier}
}

@article{spyrou2015comparative,
  title={Comparative assessment of feature extraction methods for visual odometry in wireless capsule endoscopy},
  author={Spyrou, Evaggelos and Iakovidis, Dimitris K and Niafas, Stavros and Koulaouzidis, Anastasios},
  journal={Computers in biology and medicine},
  volume={65},
  pages={297--307},
  year={2015},
  publisher={Elsevier}
}

@article{liu2013wireless,
  title={Wireless capsule endoscopy video reduction based on camera motion estimation},
  author={Liu, Hong and Pan, Ning and Lu, Heng and Song, Enmin and Wang, Qian and Hung, Chih-Cheng},
  journal={Journal of digital imaging},
  volume={26},
  number={2},
  pages={287--301},
  year={2013},
  publisher={Springer}
}

@inproceedings{werner2023precise,
  title={Precise localization within the gi tract by combining classification of cnns and time-series analysis of hmms},
  author={Werner, Julia and Gerum, Christoph and Reiber, Moritz and Nick, J{\"o}rg and Bringmann, Oliver},
  booktitle={International Workshop on Machine Learning in Medical Imaging},
  pages={174--183},
  year={2023},
  organization={Springer}
}

@article{wang2004image,
  title={Image quality assessment: from error visibility to structural similarity},
  author={Wang, Zhou and Bovik, Alan C and Sheikh, Hamid R and Simoncelli, Eero P},
  journal={IEEE transactions on image processing},
  volume={13},
  number={4},
  pages={600--612},
  year={2004},
  publisher={IEEE}
}

@article{ali2025recent,
  title={Recent advancements in localization technologies for wireless capsule endoscopy: A technical review},
  author={Ali, Muhammad A and Tom, Neil and Alsunaydih, Fahad N and Yuce, Mehmet R},
  journal={Sensors},
  volume={25},
  number={1},
  pages={253},
  year={2025},
  publisher={MDPI}
}

@article{mcdonald2009ridge,
  title={Ridge regression},
  author={McDonald, Gary C},
  journal={Wiley Interdisciplinary Reviews: Computational Statistics},
  volume={1},
  number={1},
  pages={93--100},
  year={2009},
  publisher={Wiley Online Library}
}

@book{sheather2009modern,
  title={A modern approach to regression with R},
  author={Sheather, Simon},
  year={2009},
  publisher={Springer Science \& Business Media}
}

\end{document}